\documentclass[10pt,twocolumn,letterpaper]{article}

\usepackage[accsupp]{axessibility}
\usepackage{cvpr}
\usepackage{times}
\usepackage{epsfig}
\usepackage{graphicx}
\usepackage{amsmath}
\usepackage{amssymb}
\usepackage{multirow}
\usepackage{bbold}
\usepackage[table,xcdraw]{xcolor}
\usepackage{threeparttable}
\usepackage{array}
\usepackage{bm}
\usepackage{booktabs}
\usepackage{url}
\newcommand{\PreserveBackslash}[1]{\let\temp=\\#1\let\\=\temp}
\newcolumntype{C}[1]{>{\PreserveBackslash\centering}p{#1}}
\newcolumntype{R}[1]{>{\PreserveBackslash\raggedleft}p{#1}}
\newcolumntype{L}[1]{>{\PreserveBackslash\raggedright}p{#1}}

\newcommand{\secref}[1]{Section~\ref{sec:#1}}
\newcommand{\figref}[1]{Figure~\ref{fig:#1}}
\newcommand{\tabref}[1]{Table~\ref{tab:#1}}
\newcommand{\eqnref}[1]{Eq.~\eqref{eq:#1}}

\newcommand{\mb}[1]{\mathbf{#1}}

\newlength\secmargin
\newlength\subsecmargin
\newlength\paramargin
\newlength\figmargin
\newlength\eqmargin
\newenvironment{myitemize2}[1][]{
\begin{list}{$\bullet$}
    {
     \setlength{\leftmargin}{5mm} 
     \setlength{\parsep}{0.5mm} 
     \setlength{\topsep}{0mm} 
     \setlength{\itemsep}{0mm} 
     \setlength{\labelsep}{0.5em} 
     \setlength{\itemindent}{0mm} 
     \setlength{\listparindent}{6mm} 
    }}
{\end{list}}

\usepackage[pagebackref=true,breaklinks=true,letterpaper=true,colorlinks,bookmarks=false]{hyperref}

\usepackage[capitalize]{cleveref}
\crefname{section}{Sec.}{Secs.}
\Crefname{section}{Section}{Sections}
\Crefname{table}{Table}{Tables}
\crefname{table}{Tab.}{Tabs.}

\def\cvprPaperID{3142} 
\def\confName{CVPR}
\def\confYear{2022}

\begin{document}

\title{Fine-grained Temporal Contrastive Learning for Weakly-supervised Temporal Action Localization}

\author{Junyu Gao$^{1,2}$, Mengyuan Chen$^{1,2}$, and Changsheng Xu$^{1,2,3}$\\
	$^1$ National Lab of Pattern Recognition (NLPR), \\
	Institute of Automation, Chinese Academy of Sciences (CASIA)\\
	$^2$ School of Artificial Intelligence, University of Chinese Academy of Sciences (UCAS)\\
	$^3$ Peng Cheng Laboratory, ShenZhen, China \\
	{\tt\small {\{junyu.gao, csxu\}}@nlpr.ia.ac.cn; chenmengyuan2021@ia.ac.cn}
}

\maketitle

\begin{abstract}
\vspace{\secmargin}
We target at the task of weakly-supervised action localization (WSAL), where only video-level action labels are available during model training. 
Despite the recent progress, existing methods mainly embrace a localization-by-classification paradigm and overlook the fruitful fine-grained temporal distinctions between video sequences, thus suffering from severe ambiguity in classification learning and classification-to-localization adaption. This paper argues that learning by contextually comparing sequence-to-sequence distinctions offers an essential inductive bias in WSAL and helps identify coherent action instances. Specifically, under a differentiable dynamic programming formulation, two complementary contrastive objectives are designed, including Fine-grained Sequence Distance (FSD) contrasting and Longest Common Subsequence (LCS) contrasting, where the first one considers the relations of various action/background proposals by using match, insert, and delete operators and the second one mines the longest common subsequences between two videos. Both contrasting modules can enhance each other and jointly enjoy the merits of discriminative action-background separation and alleviated task gap between classification and localization. Extensive experiments show that our method achieves state-of-the-art performance on two popular benchmarks. Our code is available at \url{https://github.com/MengyuanChen21/CVPR2022-FTCL}.

\end{abstract}

\vspace{-4mm}
\section{Introduction}\label{sec:intro}
\vspace{\secmargin}
Action localization is one of the most fundamental tasks in computer vision, which aims to localize the start and end timestamps of different actions in an untrimmed video~\cite{zhu2021enriching,sridhar2021class,long2019gaussian,xu2020gtad}. 
In the past few years, the performance has gone through a phenomenal surge under the fully-supervised setting. However, collecting and annotating precise frame-wise information is a bottleneck and consequently limits the scalability of a fully supervised framework for real-world scenarios. Therefore, weakly-supervised action localization (WSAL) has been explored~\cite{hong2021cross,huang2021foreground,qu_2021_acmnet,yang2021uncertainty}, where only video-level category labels are available.

\begin{figure}[t]
	\centering
	\includegraphics[width=0.95\linewidth]{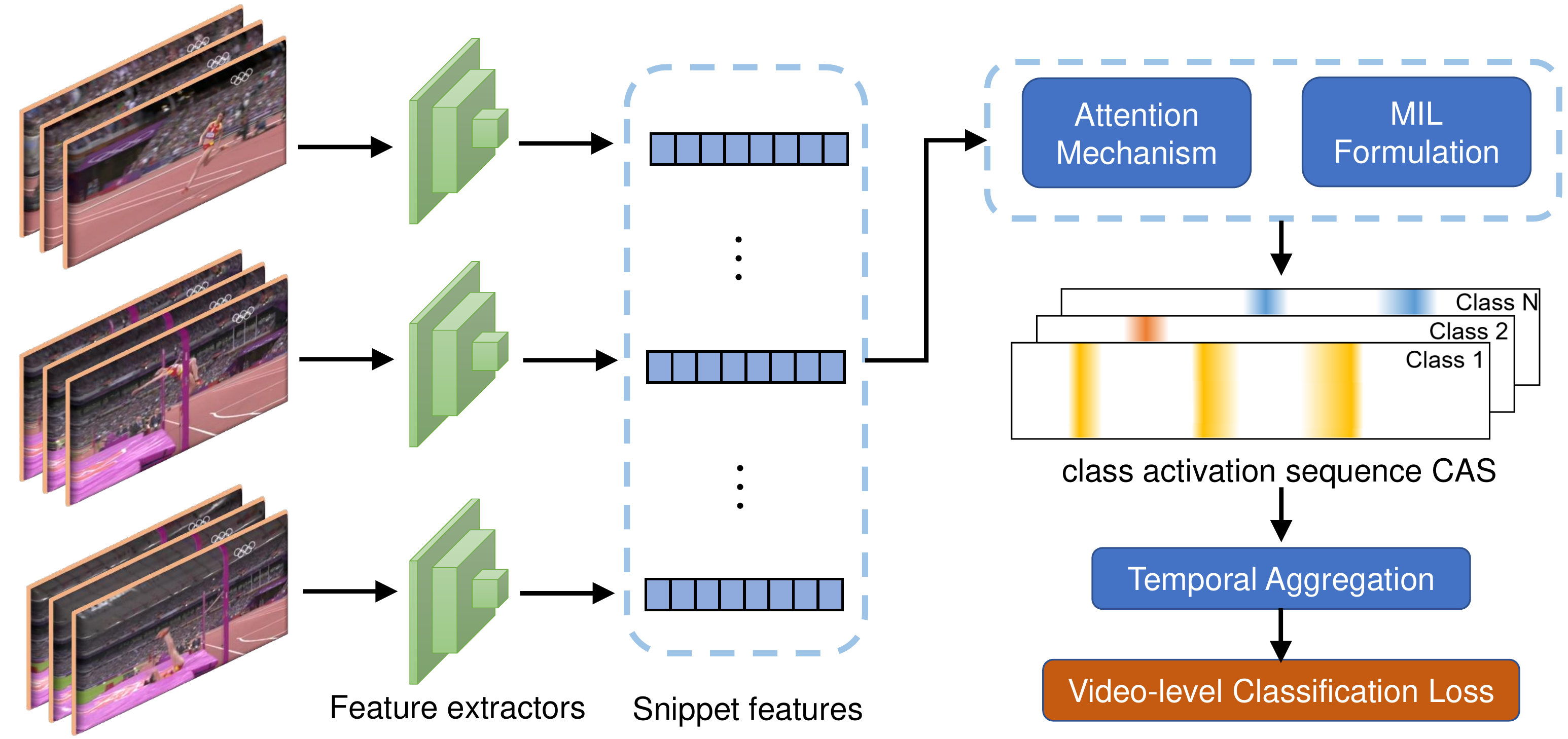}
	\vspace{-2mm}
	\caption{Pipeline of the localization-by-classification paradigm. It first extracts snippet-level features and adopts attention/MIL mechanisms for learning CAS under video-level supervisions.
	}\label{fig:motivation-pipeline}
	\vspace{-3mm}
\end{figure}

To date in the literature, current approaches mainly embrace a localization-by-classification paradigm~\cite{shi2020weakly,xu2019segregated,paul2018wtalc,wang2017untrimmednets}, which divides each input video into a series of fixed-size non-overlapping snippets and aims for generating the temporal Class Activation Sequences (CAS)~\cite{zhang2021cola,qu_2021_acmnet}.  Specifically, as shown in~\figref{motivation-pipeline}, by optimizing a video-level classification loss, most existing WSAL approaches adopt the multiple instance learning (MIL) formulation~\cite{ma2021weakly} and attention mechanism~\cite{qu_2021_acmnet} to train models to assign snippets with different class activations. The final action localization results are inferred by thresholding and merging these activations. To improve the accuracy of learned CAS, various strategies have been proposed, such as uncertainty modeling~\cite{yang2021uncertainty}, collaborative learning~\cite{hong2021cross, huang2021foreground}, action unit memory~\cite{luo2021actionunit}, and causal analysis~\cite{ liu2021blessings }, which have obtained promising performance.

\begin{figure*}[t]
	\centering
	\includegraphics[width=0.95\linewidth]{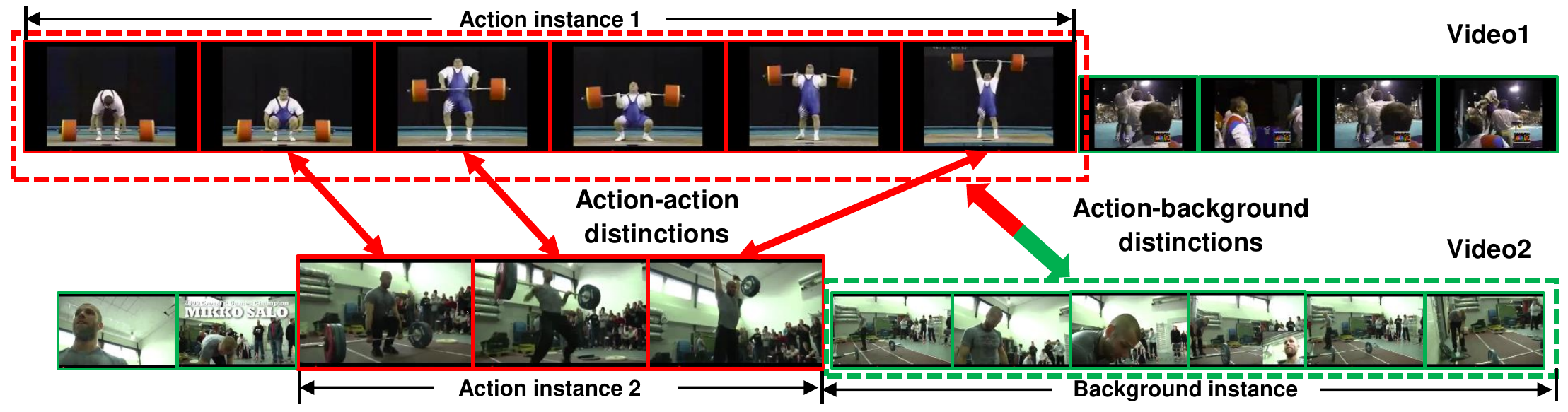}
	\vspace{-2mm}
	\caption{ Fine-grained temporal distinctions between two videos. Here, the two untrimmed videos are from the same action category \emph{CleanAndJerk}. Note that the distinctions are derived from two aspects: (1) Fine-grained action-background distinctions. The snippets in the action instances and background subsequences are semantically different, which should be effectively separated in a robust WSAL model. (2) Fine-grained distinctions between action instances. In this example, the three snippets of the action instance in Video2 can be aligned with the partial action instance in Video1. In addition, we can observe that the three snippets linked by the red arrows are the longest common sequences of both videos. We argue that considering the above fine-grained distinctions can benefit WSAL learning.
	}\label{fig:motivation-ft}
	\vspace{-3mm}
\end{figure*}

Despite achieving significant progress, the above learning pipelines still suffer from severe localization ambiguity due to the lack of fine-grained frame-wise annotations in the temporal dimension, which dramatically hinders the WSAL performance of the localization-by-classification paradigm.    
Specifically, the ambiguity is two-fold: \textbf{(1)} Without sufficient annotations in the weakly-supervised setting, the learned classifier itself is not discriminative and robust enough, causing difficulties in action-background separation. \textbf{(2)} Since there exists a large task gap between classification and localization, the learned classifiers usually focus on the easy-to-distinguish snippets while ignoring those that are not prominent in localization. As a result, the localized temporal sequences are often incomplete and inexact. 

To alleviate the above ambiguity, we argue that videos naturally provide a rich source of temporal structures and additional constraints for improving weakly-supervised learning. As outlined in~\figref{motivation-ft}, an action video generally includes a series of fine-grained snippets, while different action/background instances possess correlative and fine-grained temporal distinctions. For example, given a pair of videos from the same action category but captured in varied scenes, there exists a latent temporal association between both videos. With this in mind, a key consideration is to leverage such temporal distinctions for improving representation learning in WSAL.
However, when elaborately comparing two videos, no guarantee ensures that they can be aligned directly. Recently, dynamic time warping (DTW)~\cite{berndt1994using, sakoe1978dynamic} was proposed to tackle the misalignment issue in various video analysis tasks such as action classification~\cite{hadji2021representation}, few-shot learning~\cite{cao2020few}, action segmentation and video summarization~\cite{chang2021learning, chang2019d3tw}. DTW computes the discrepancy between two videos based on their optimal alignment from dynamic programming.  However, the above approaches either assume the video is trimmed~\cite{hadji2021representation,cao2020few} or require additional supervision~\cite{chang2021learning,chang2019d3tw} such as action orders, which impedes the direct use of DTW in WSAL. 

In this paper, to address the above issues, we propose a novel Fine-grained Temporal Contrastive Learning (FTCL) framework for weakly-supervised temporal action localization. By capturing the distinctive temporal dynamics of different video sequences, FTCL focuses on optimizing the structural and fine-grained snippet-wise relations between videos by leveraging end-to-end differentiable dynamic programming goals, with loss that is informed from the structural relations. Specifically, \textbf{(1)} To improve the robustness of action-background separation, we contrast the fine-grained sequence distance (FSD) calculated from different action/background instance pairs by designing an improved and differentiable edit distance measurement. The measurement can evaluate whether two sequences are structurally analogous by calculating the minimum cost required to transform one to the other. \textbf{(2)} To alleviate the task gap between classification and localization, we aim at contrasting the mined Longest Common Subsequence (LCS) between two untrimmed videos that contain the same action. Different video sequences from the same category can provide complementary clues for exploring the complete action instance by optimizing the LCS. Therefore, LCS learning between different video sequences improves the coherence in a predicted action instance. Finally, with FSD and LCS contrasting, a unified framework is constructed in an end-to-end manner, while the proposed FTCL strategy can be seamlessly integrated into any existing WSAL approach.

The main contributions of this paper are three-fold:
\vspace{\figmargin}
\begin{myitemize2}
	\item 
	In light of the above analysis, we contend that localizing action by contextually contrasting fine-grained temporal distinctions offers an essential inductive bias in WSAL. We thus introduce the first discriminative sequence-to-sequence comparing framework for robust WSAL to address the lack of frame-wise annotations, capable of leveraging fine-grained temporal distinctions.
	
	\item A unified and differentiable dynamic programming formulation, including fine-grained sequence distance learning and longest common subsequence mining, is designed, which jointly enjoys the merits of (1) discriminative action-background separation and (2) alleviated task gap between classification and localization.
	
	\item Extensive experimental results on two popular benchmarks demonstrate that the proposed FTCL algorithm performs favorably. Note that the proposed strategy is model-agnostic and non-intrusive, and hence can play a complementary role over existing methods to promote the action localization performance consistently.
\end{myitemize2}

\vspace{\secmargin}
\section{Related Work}\label{sec:related_work}
\vspace{\secmargin}
\noindent\textbf{Fully-supervised Temporal Action Localization (TAL).}
Compared with traditional video understanding tasks~\cite{carreira2017quo,gao2018watch,gao2019graph,gao2021fast,gao2017deep_relative_tracking}, TAL aims to classify every activity instance in an untrimmed video and predict their accurate temporal locations. Existing TAL approaches can be roughly divided into two categories: two-stage methods~\cite{zhu2021enriching,sridhar2021class,chao2018rethinking,shou2016temporal,xu2017rc3d,zhao2017temporal,dai2017temporal} and one-stage methods~\cite{lin2021learning,tan2021relaxed,buch2017SSTAD,lin2017single,shou2017cdc,long2019gaussian,xu2020gtad}. For the former one, action proposals are firstly generated and then fed into a classifier. This pipeline mainly focuses on improving the quality of proposals~\cite{zhu2021enriching,shou2016temporal,chao2018rethinking} and the robustness of classifiers~\cite{sridhar2021class,zhao2017temporal}. 
One-stage methods instead predict action location and category simultaneously. SS-TAD~\cite{buch2017SSTAD} utilizes recurrent neural networks to regress the temporal boundaries and action labels jointly. Lin~\etal~\cite{lin2021learning} introduces an anchor-free framework in a coarse-to-fine manner. Although the above model achieves significant performance, the fully-supervised setting limits their scalability and practicability in the real world~\cite{junyu2019AAAI_TS-GCN,gao2020learning,gao2021learningUVMR}.

\noindent\textbf{Weakly-supervised Action Localization.}
To overcome the above limitation, WSAL has drawn significant attention in recent years by leveraging different types of supervisions, \eg, web videos~\cite{gan2016webly}, action orders~\cite{bojanowski2014weakly}, single-frame annotation~\cite{ma2020sf,lee2021learningSP}, and video-level category labels~\cite{wang2017untrimmednets,nguyen2018weakly,liu2019completeness}. Among these weak supervisions, the last one is the most commonly used due to the low cost. UntrimmedNet~\cite{wang2017untrimmednets} is the first work that uses video-level category labels for WSAL via a relevant segment selection module. Currently, most existing approaches can be roughly divided into three groups, namely attention-based methods~\cite{shi2020weakly,xu2019segregated,hong2021cross,qu_2021_acmnet,hong2021cross,narayan2021d2,luo2021actionunit,liu2019weaklyICCV}, MIL-based methods~\cite{ma2021weakly,lee2020background,luo2020weakly,moniruzzaman2020action,paul2018wtalc}, and erasing-based methods~\cite{singh2017hide,zhang2019adversarial,zhong2018step}. Attention-based approaches aim at selecting snippets of high activation scores and suppressing background snippets. ACM-Net~\cite{qu_2021_acmnet} investigates a three-branch attention module by simultaneously and effectively considering action instances, context, and background information. MIL-based pipeline treats the entire video as a bag and utilizes a top-$k$ operation to select positive instances. W-TALC~\cite{paul2018wtalc} introduces a co-activity relation loss to model inter- and intra-class information. The erasing-based methods, \eg, Hide-and-Seek~\cite{singh2017hide}, typically attempt to erase input segments during training for highlighting less discriminative snippets. 

Note that most existing methods only consider the video-level supervision but ignore the fine-grained temporal distinctions between videos, and can hardly benefit from discriminative learning of snippet-wise contrasting. Although some approaches have investigated different types of contrastive regularization, \eg, hard snippet contrasting in CoLA~\cite{zhang2021cola}, they perform contrasting by only considering video-level information ~\cite{paul2018wtalc,narayan20193c,islam2020weakly} or neglecting the fine-grained temporal structures~\cite{zhang2021cola,narayan2021d2,nguyen2019weakly}.  To the best of our knowledge, we are the first to introduce the contrastive learning of fine-grained temporal distinctions to the WSAL task. Experimental results demonstrate that the proposed FTCL learns discriminative representations, thus facilitating the action localization.

\noindent\textbf{Dynamic Programming for Video Understanding.}
Recent progress has shown that learning continuous relaxation of discrete operations (\eg, dynamic programming) can benefit video representation learning~\cite{hadji2021representation,cao2020few,chang2021learning, chang2019d3tw}. A popular framework is to adopt sequence alignment as a proxy task and then uses dynamic time warping (DTW) to find the optimal alignment~\cite{berndt1994using, sakoe1978dynamic,cai2019dtwnet,mensch2018differentiable,cuturi2017soft,dvornik2021drop,dwibedi2019temporal}. For example, based on a novel probabilistic path finding view, Hadji~\etal~\cite{hadji2021representation} design contrastive and cycle-consistency objectives for video representation learning by leveraging differentiable DTW. Chang~\etal~\cite{chang2021learning} propose discriminative prototype DTW to learn class-specific prototypes for temporal action recognition. However, the above dynamic programming strategies either assume the video is trimmed~\cite{hadji2021representation,cao2020few} or require additional supervision~\cite{chang2021learning,chang2019d3tw} such as action orders, thus cannot be applied to the WSAL task. Different from the above approaches, this paper proposes to leverage fine-grained sequence distance and longest common subsequence contrasting for discriminative foreground-background separation and robust classification-to-localization adaption.

\section{Our Approach}\label{sec:our}
\begin{figure}[t]
	\centering
	\includegraphics[width=1\linewidth]{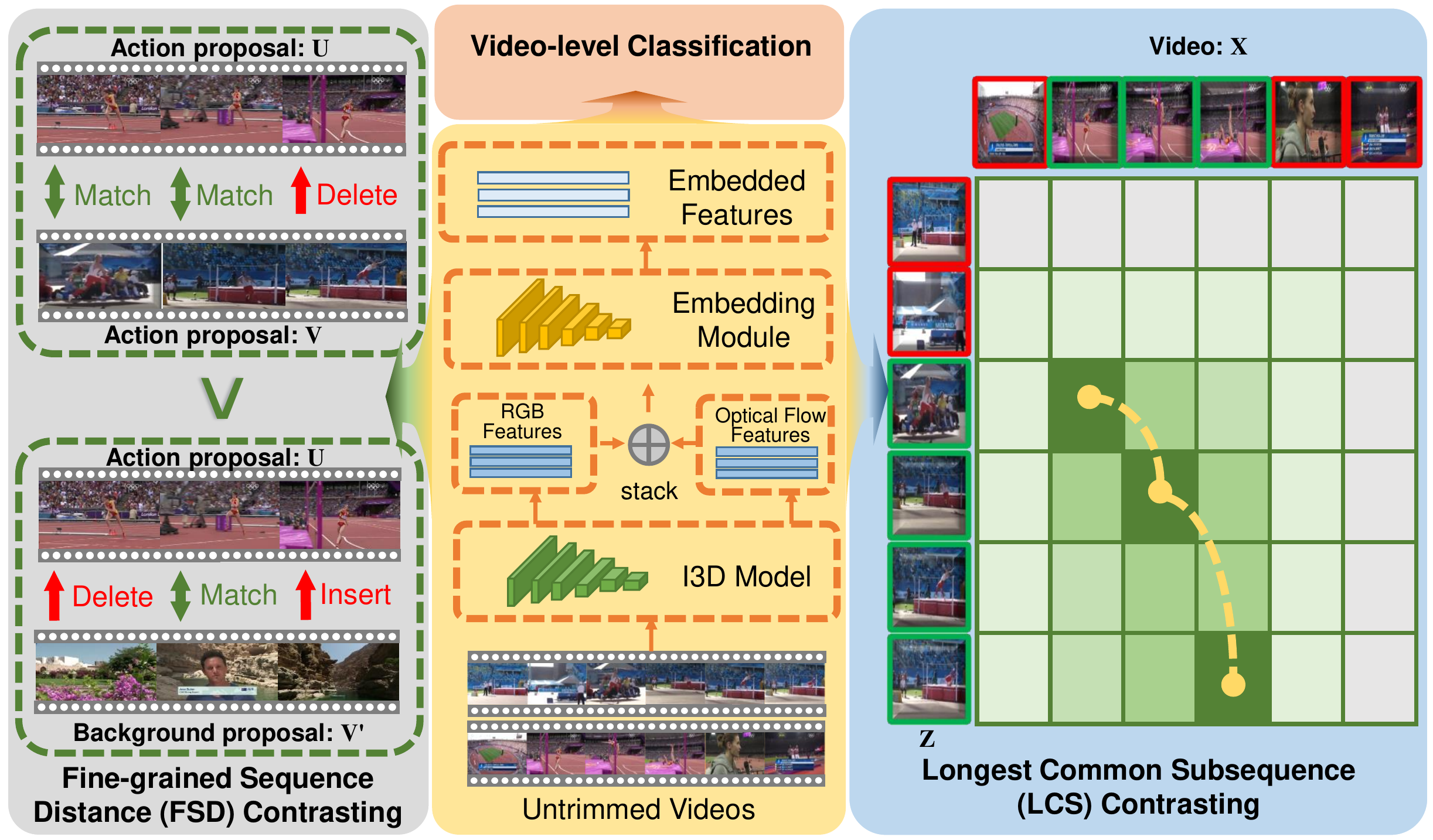}
	\vspace{-3mm}
	\caption{Our proposed FTCL architecture and toy example. The pretrained I3D model is first adopted for the input videos to obtain RGB and optical flow features. Then an embedding module is further applied to extract snippet-wise features under video-level supervisions. To achieve discriminative action-background separation, FSD contrasting is designed to consider the relations of different action/background proposals using Match, Insert, and Delete operators. For classification-to-localization adaption, we employ LCS contrasting to find the longest common subsequences between two videos. Both contrasting strategies are implemented via differentiable dynamic programming.
	}\label{fig:arch}
	\vspace{-3mm}
\end{figure}

\vspace{-1mm}
In this work, we describe our WSAL approach based on Fine-grained Temporal Contrastive Learning (FTCL). As shown in~\figref{arch}, given a set of video sequence pairs, our training objective is the learning of an embedding function applied to each snippet. We firstly adopt feature extractors to obtain the appearance (RGB) and motion (optical flow) features of each snippet (\secref{our-basic}). Then, under a differentiable dynamic programming formulation, two complementary contrastive objectives are designed for learning fine-grained temporal distinctions including Fine-grained Sequence Distance (FSD) contrasting (\secref{our-fsd}) and Longest Common Subsequence (LCS) contrasting (\secref{our-lcs}). Finally, the whole framework is end-to-end learned (\secref{our-loss-infer}), which can jointly achieve discriminative action-background separation and alleviated task gap between classification and localization.

\subsection{Notations and Preliminaries}\label{sec:our-basic}
Given an untrimmed video $\mb X$ with its groundtruth label $\mb y \in \mathbb{R}^C$, where $C$ is the number of action categories. $\mb y_i = 1$ if the $i$-th action class is present in the video and $\mb y_i = 0$ otherwise. For the video, we divide it into non-overlapping $T$ snippets and apply feature extractors to obtain snippet-wise features $\mb X = [\mb x_1,...,\mb x_i,..., \mb x_T] \in \mathbb{R}^{D \times T}$, where $D$ is the feature dimension and each snippet has 16 frames. In this paper, for a fair comparison, we follow previous approaches~\cite{narayan20193c,paul2018wtalc,qu_2021_acmnet,zhang2021cola} to extract features from both RGB and optical flow streams by using the I3D network~\cite{carreira2017quo} pretrained on the Kinetics dataset. After that, the two types of features are concatenated together and then input into an embedding module, \eg, convolutional layers~\cite{qu_2021_acmnet}, for generating $\mb X$. The goal of WSAL is to learn a model that simultaneously localizes and classifies all action instances in a video with timestamps as $(t_s, t_e, c, \phi)$, where $t_s, t_e, c$, and $\phi$ denote the start time, the end time, the predicted action category and the confidence score of the action proposal, respectively.

Currently, existing dominant approaches mainly embrace a localization-by-classification framework, which first learns importance scores for aggregating snippet-level features into a video-level embedding and then perform action classification by using the video-level labels:
\begin{align}
	\begin{split}
		\overline{\mb{x}}&=\sum_{t=1}^{T} \alpha_{t} * \mb{x}_{t} \\ 
		\mathcal{L}_{c l s}&=-\sum_{i=1}^{C} \mb y_{i} \log \widetilde{\mb y}_{i}
	\end{split}
\end{align}
where $\alpha_{t} = f_{\alpha}(\mb x_t)$ is the learned importance score. The generated video-level feature is further fed into a classifer to obtain the prediction results $\widetilde{\mb y} = f_{cls}(\overline{\mb{x}})$. After model training, $f_{\alpha}(\cdot)$ and $f_{cls}(\cdot)$ is used for inferring the snippet-level Class Activation Sequences (CAS) of a test video. To learn the two functions, various strategies can be applied such as multiple attention learning~\cite{qu_2021_acmnet} and modality collaborative learning~\cite{hong2021cross}.

\subsection{Discriminative Aciton-Background Separation via FSD Contrasting}\label{sec:our-fsd}
To learn discriminative action-background separation in the above localization-by-classification framework, a few existing methods resort to performing contrastive learning by either using global video features~\cite{paul2018wtalc,narayan20193c,islam2020weakly} or only considering intra-video contrast without temporal modeling~\cite{zhang2021cola,narayan2021d2,nguyen2019weakly}.  However, these models ignore the fine-grained temporal distinctions between videos, resulting in the insufficient discriminative ability for classification. 

In this work, we propose to contrast two video sequences temporally in a fine-grained manner. Existing methods usually calculate the similarity of two sequences by measuring the vector distance between their global feature representations. Different from this matching strategy, as shown in the left of~\figref{arch}, we would like to determine whether two sequences are structurally analogous by evaluating the minimum cost required to transform one sequence to the other. The naive idea is to exhaustively compare all the possible transformations, which is NP-hard. A fast solution is to utilize solvable dynamic programming techniques, where sub-problems can be nested recursively inside larger problems. Here, motivated by the widely used edit distance\footnote{Edit distance is a way of quantifying how dissimilar two strings are to one another by counting the minimum number of operations required to transform one string into the other.}~\cite{navarro2001guided} in computational linguistics and computer science, we design differentiable \emph{Match}, \emph{Insert}, and \emph{Delete} operators for sequence-to-sequence similarity calculation. Specifically, with the learned CAS, we can generate various action/background proposals, where an action proposal $\mb U$ contains snippets with high action activations and a background proposal $\mb V$ is just the opposite.  For the two proposal sequences with lengths of $M$ and $N$, $\mb U = [\mb u_1,...,\mb u_i,..., \mb u_M]  \in \mathbb{R}^{D \times M}$ and $\mb V = [\mb v_1,...,\mb v_i,..., \mb v_M] \in \mathbb{R}^{D \times N}$, their similarity is evaluated with the following recursion:

\begin{equation}\label{eq:fsd}
	\mb S(i, j)= \mu_{i,j} + \max 
	\begin{cases}
		\mb S(i-1, j-1) & (\text { Match }) \\ 
		g_{i, j}+\mb S(i-1, j) & (\text { Insert }) \\ 
		h_{i, j}+\mb S(i, j-1) & (\text { Delete })
	\end{cases}
\end{equation}
where the sub-sequence similarity score $\mb S(i, j)$ is evaluated on position $i$ in the first sequence $\mb U$ and on position $j$ in the second sequence $\mb V$. $\mb S(0, :)$ and $\mb S(:, 0)$ are initialized to zeros. Intuitively, in position $(i,j)$, if $\mb u_i$ and $\mb v_j$ are matched, the sequence similarity score should be increased. If the insert or delete operation is conducted, there should be a penalty on the similarity score. To this end, we learn three types of residual values (scalars),  $\mu_{i,j}$, $g_{i,j}$, and $h_{i,j}$ for these operations. Taking $\mu_{i,j}$ and $g_{i,j}$ as an example, which can be calculated as follows:
\begin{equation}\label{eq:mug}
	\mu_{i,j} = \sigma_{\mu} (\text{cos}(\bm \Delta_{i,j}^{\mu})), \quad
	g_{i,j} = \sigma_{g} (\text{cos}(\bm \Delta_{i,j}^{g}))
\end{equation}
where $\bm \Delta_{i,j}^{\mu} = [f_{\mu}(\mb u_i), f_{\mu}(\mb v_j)]$ and  $\bm \Delta_{i,j}^{g}$ is defined similarly. $f_{\mu}(\cdot)$, $f_{g}(\cdot)$, and $f_{h}(\cdot)$ are three fully-connected layers. We utilize these functions to simulate different operations including match, insert, and delete.
$\sigma_{\mu}$ and $\sigma_{g}$ are activation functions for obtaining the residual values. 

After conducting the above recursive calculation, $\mb S(i, j)$ is guaranteed to be the optimal similarity score between the two sequences. It is evident that the similarity between two action proposals from the same category should be larger than it between an action proposal and a background proposal. By leveraging this relation, we design the FSD contrasting loss as follows:

\begin{equation}\label{eq:fsd-loss}
	\mathcal{L}_{\mathrm{FSD}}=\ell\left(s_{\left[\mb U \mb V^{\prime} \right]}-s_{[\mb U \mb V]}\right)+\ell\left(s_{\left[\mb U^{\prime} \mb V \right]}-s_{[\mb U \mb V]}\right)
\end{equation}
where $\ell(x)$ denotes the ranking loss. The subscript $[\mb U \mb V]$ indicates the two action proposals from the same category for calculating the sequence-to-sequence similarity $s = \mb S(M,N)$. $\mb U^{\prime}$ and $\mb V^{\prime}$ represents the background proposals. In our implementation, we utilize the learned importance score $\alpha$~\cite{qu_2021_acmnet} to select action and background proposals.

\noindent\textbf{Smooth Max Operation.}
As the max operation in~\eqnref{fsd} is not differentiable, the recursive matrices and the traceback cannot be differentiated in current formulation. Therefore, we are motivated to utilize a standard smooth approximation for the max operator~\cite{mensch2018differentiable}:
\begin{equation}\label{eq:smoothMax}
	\text{smoothMax}(\mb a; \gamma) = \log(\sum_i \exp(\gamma \mb a_i))
\end{equation}
where $\mb a = [\mb a_1,...,\mb a_i,...]$ is a vector for max operator. $\gamma$ represents the temperature hyper-parameter. Note that other types of smooth approximation~\cite{cuturi2017soft,cai2019dtwnet,hadji2021representation} can also be applied for differentiating while designing a novel smooth max operation is not the goal of our paper.

\subsection{Robust Classification-to-Localization Adaption via LCS Contrasting}\label{sec:our-lcs}
In the above section, action-background separation is considered, which improves the discriminative ability of the learned action classifiers. However, the goal of WSAL task is to localize action instances temporally with precise timestamps, resulting in a large task gap between classification and localization. To alleviate this gap, we attempt to mine the longest common subsequence (LCS) between two untrimmed videos $\mb X$ and $\mb Z$ thus improve the coherence in the learned action proposals. The intuition behind this idea is two-fold: (1) If the two videos do not share the same actions, the length of LCS between $\mb X$ and $\mb Z$ should be small. Obviously, due to the diverse background and substantial difference between the two types of actions, snippets from the two individual videos are likely to be highly inconsistent, resulting in short LCS. (2) Similarly, if two videos share the same action, their LCS is prone to be long since action instances from the same category are composed of similar temporal action snippets. Ideally, the LCS in this situation is as long as the shorter action instance. For example, as shown in~\figref{motivation-ft}, the action \emph{CleanAndJerk} consists of several sequential sub-actions like \emph{squat}, \emph{grasp}, and \emph{lift}. 

Based on the above observation, as shown in the right of~\figref{arch}, we propose to model the LCS between $\mb X$ and $\mb Z$ by designing a differentiable dynamic programming strategy. Specifically, we maintain a recursive  matrix $\mb R \in \mathbb{R}^{(T+1) \times (T+1)}$, with elements $\mb R(i,j)$ stores the length of longest common subsequence of prefixes $\mb X_{i}$ and $\mb Z_{j}$. To find the LCS of prefixes $\mb X_{i}$ and $\mb Z_{j}$, we first compare $\mb x_i$ and $\mb z_j$. If they are equal, then the calculated common subsequence is extended by that element and thus $\mb R(i, j) = \mb R(i-1, j-1) +1$. If they are not equal, the largest length calculated before is retained for $\mb R(i, j)$. In the WSAL task, since a pair of snippets cannot be exactly the same even they depict the same action, we adopt their similarities to calculate the accumulated soft length of two sequences. As a result, we design the recursion formula of LCS modeling:

\begin{equation}\label{eq:lcs}
	\mb R(i, j)=\left\{\begin{array}{c}
		0, \quad i=0 \text { or } j=0 \\
		\mb R(i-1, j-1)+c_{i,j}, \quad c_{i,j} \geqslant \tau  \\
		\max \{\mb R(i-1, j), \mb R(i, j-1)\}, c_{i,j} < \tau
	\end{array}\right.
\end{equation}
where $\tau$ is a threshold that determines whether the $i$-th snippet of video $\mb X$ and the $j$-th snippet of video $\mb Z$ is matched. $c_{i,j} = \text{cos}(\mb x_i, \mb z_j)$ is the cosine similarity of snippets $\mb x_i$ and $\mb z_j$. Note that by using the equation above, we can seek the longest common subsequence between two videos. Although not used here, the mined subsequence can qualitatively demonstrate the effectiveness and improve the interpretability of our approach (\secref{ablation}).

With the above dynamic programming, the resulting values $r = \mb R(T,T)$  represents the soft length of the longest common subsequence between the two videos. We utilize a cross-entropy loss to serve as a constraint for LCS learning:
\begin{equation}\label{eq:lcs-loss}
	\mathcal{L}_{\mathrm{LCS}}=\delta_{xz} \log(r_{[\mb X \mb Z]}) + (1-\delta_{xz})\log(1-r_{[\mb X \mb Z]})
\end{equation}
where $\delta_{xz}$ is the groundtruth indicating whether the two videos $\mb X$ and $\mb Z$ have the same action categories. 


\noindent\textbf{Discussion.} In this work, FSD and LCS learning strategies are proposed via differentiable dynamic programming, while both are designed for sequence-to-sequence contrasting. However, the two modules are not redundant and have substantial difference: (1) They have different goals by considering different types of sequences. We utilize FSD to learn robust action-background separation while different action and background proposals are employed. While LCS contrasting is designed to find coherent action instances in two untrimmed videos, thus achieving classification-to-localization adaption. (2) They have different contrasting levels. In FSD contrasting, the relations between different action/background pairs are considered~(\eqnref{fsd-loss}), whereas in LCS, the contrasting is conducted in a pair of untrimmed videos~(\eqnref{lcs-loss}). 
We also demonstrate that jointly learning FSD and LCS can enhance and complement each other for pursuing effective WSAL in~\secref{ablation}.

\subsection{Learning and Inference}\label{sec:our-loss-infer}
\noindent\textbf{Training.} 
The above two objectives can be seamlessly integrated into existing WSAL frameworks and collaborate with each other.
For optimizing the whole model, we compose the classification loss and the two contrastive losses:
\begin{equation} 
	\mathcal{L} =  \mathcal{L}_{c l s}+ \mathcal{L}_{\mathrm{FSD}}+\mathcal{L}_{\mathrm{LCS}}
\end{equation}
Since our proposed method is model-agnostic and non-intrusive, the two contrastive losses can well cooperate with any other weakly-supervised action localization objectives by replacing $\mathcal{L}_{c l s}$ with different types of loss functions and backbones (Please refer to~\secref{ablation}).

\noindent\textbf{Inference.} 
Given a  test video, we first predict the snippet-level CAS and then apply a threshold strategy to obtain action snippet candidates following the standard process~\cite{qu_2021_acmnet}. Finally, continuous snippets are grouped into action proposals, and then we perform non-maximum-suppression (NMS) to remove duplicated proposals.


\section{Experimental Results}\label{sec:experimental_results}
\vspace{\secmargin}

We evaluate the proposed FTCL on two popular datasets: THUMOS14~\cite{idrees2017thumos} and ActivityNet1.3~\cite{caba2015activitynet}. Extensive experimental results demonstrate the effectiveness of our proposed method.


\begin{table*}[]
	\centering
	\caption{Temporal action localization performance comparison with state-of-the-art methods on the THUMOS14 dataset. Note that weak$^+$ represents methods that utilize external supervision information besides video labels.}
	\vspace{-4mm}
	\label{tab:THUMOS}
	\small
	\resizebox{160mm}{!}{%
	\renewcommand{\arraystretch}{1.1} 
	\begin{tabular}{@{}c|c|cccccccccc@{}}
		\toprule
		\multirow{2}{*}{Supervision} & \multirow{2}{*}{Method} & \multicolumn{10}{c}{mAP@t-IoU(\%)}                                                                                                                                                                \\
		&                         & 0.1           & 0.2           & 0.3           & 0.4           & 0.5           & 0.6           & 0.7           & {[}0.1:0.5{]} & \multicolumn{1}{l}{{[}0.3:0.7{]}} & \multicolumn{1}{l}{Avg} \\ \midrule
		\multirow{6}{*}{Fully}       & S-CNN\cite{shou2016action}, CVPR2016         & 47.7          & 43.5          & 36.3          & 28.7          & 19.0          & -             & -             & 35.0             & -                                    & -                       \\
		& CDC\cite{shou2017cdc}, CVPR2017           & -             & -             & 40.1          & 29.4          & 23.3          & 13.1          & 7.9           & -                & -                                    & -                       \\
		& R-C3D\cite{xu2017rc3d}, ICCV2017         & 54.5          & 51.5          & 44.8          & 35.6          & 28.9          & -             & -             & 43.1             & -                                    & -                       \\
		& SSN\cite{zhao2017temporal}, ICCV2017           & 66.0          & 59.4          & 51.9          & 41.0          & 29.8          & -             & -             & 49.6             & -                                    & -                       \\
		& TAL-Net\cite{chao2018rethinking}, CVPR2018       & 59.8          & 57.1          & 53.2          & 48.5          & 42.8          & 33.8          & 20.8          & 52.3             & 39.8                                 & 45.1                    \\
		& GTAN\cite{long2019gaussian}, CVPR2019          & 69.1          & 63.7          & 57.8          & 47.2          & 38.8          & -             & -             & 55.3             & -                                    & -                       \\ \midrule
		\multirow{2}{*}{Weakly$^+$}    & STAR\cite{xu2019segregated}, AAAI2019          & 68.8          & 60.0          & 48.7          & 34.7          & 23.0          & -             & -             & 47.0             & -                                    & -                       \\
		& 3C-Net\cite{narayan20193c}, ICCV2019        & 59.1          & 53.5          & 44.2          & 34.1          & 26.6          & -             & 8.1           & 43.5             & -                                    & 37.6                    \\ \midrule
		\multirow{22}{*}{Weakly}       & UntrimmedNet\cite{wang2017untrimmednets}, CVPR2017  & 44.4          & 37.7          & 28.2          & 21.1          & 13.7          & -             & -             & 29.0             & -                                    & -                       \\
		& Hide-and-Seek\cite{singh2017hide}, ICCV2017 & 36.4          & 27.8          & 19.5          & 12.7          & 6.8           & -             & -             & 20.6             & -                                    & -                       \\
		& AutoLoc\cite{shou2018autoloc}, ECCV2018       & -             & -             & 35.8          & 29.0          & 21.2          & 13.4          & 5.8           & -                & -                                    & -                       \\
		& STPN\cite{nguyen2018weakly}, CVPR2018          & 52.0          & 44.7          & 35.5          & 25.8          & 16.9          & 9.9           & 4.3           & 35.0             & 18.5                                 & 27.0                    \\
		& W-TALC\cite{paul2018wtalc}, ECCV2018         & 55.2          & 49.6          & 40.1          & 31.1          & 22.8          & -             & 7.6           & 39.8             & 25.4                                 & 34.4                    \\
		& CMCS\cite{liu2019completeness}, CVPR2019          & 57.4          & 50.8          & 41.2          & 32.1          & 23.1          & 15.0          & 7.0           & 40.9             & 23.7                                 & 32.4                    \\
		& WSAL-BM\cite{nguyen2019weakly}, ICCV2019       & 60.4          & 56.0          & 46.6          & 37.5          & 26.8          & 19.6          & 9.0           & 45.5             & 27.9                                 & 36.6                    \\
		& DGAM\cite{shi2020weakly}, CVPR2020          & 60.0          & 54.2          & 46.8          & 38.2          & 28.8          & 19.8          & 11.4          & 45.6             & 29.0                                 & 37.0                    \\
		& TCAM\cite{gong2020learning}, CVPR2020          & -             & -             & 46.9          & 38.9          & 30.1          & 19.8          & 10.4          & -                & 29.2                                 & -                       \\
		& Bas-Net\cite{lee2020background}, AAAI2020       & 58.2          & 52.3          & 44.6          & 36.0          & 27.0          & 18.6          & 10.4          & 43.6             & 27.3                                 & 35.3                    \\
		& A2CL-PT\cite{min2020adversarial}, ECCV2020       & 61.2          & 56.1          & 48.1          & 39.0          & 30.1          & 19.2          & 10.6          & 46.9             & 29.4                                 & 37.8                    \\
		& RefineLoc\cite{alwassel2019refineloc}, WACV2021     & -             & -             & 40.8          & 32.7          & 23.1          & 13.3          & 5.3           & -                & 23.0                                 & -                       \\
		& Liu et al\cite{liu2021weakly}, AAAI2021    & -             & -             & 50.8          & 41.7          & 29.6          & 20.1          & 10.7          & -                & 30.6                                 & -                       \\
		& ACSNet\cite{liu2021acsnet}, AAAI2021        & -             & -             & 51.4          & 42.7          & 32.4          & 22.0          & 11.7          & -                & 32.0                                 & -                       \\
		& HAM-Net\cite{islam2021hybrid}, AAAI2021       & 65.9          & 59.6          & 52.2          & 43.1          & 32.6          & 21.9          & 12.5          & 50.7             & 32.5                                 & 41.1                    \\
		& Lee et al\cite{lee2020weakly}, AAAI2021    & 67.5          & 61.2          & 52.3          & 43.4          & 33.7          & 22.9          & 12.1          & 51.6             & 32.9                                 & 41.9                    \\
		& ASL\cite{ma2021weakly}, CVPR2021           & 67.0          & -             & 51.8          & -             & 31.1          & -             & 11.4          & -                & -                                    & 40.3                    \\
		& CoLA\cite{zhang2021cola}, CVPR2021          & 66.2          & 59.5          & 51.5          & 41.9          & 32.2          & 22.0          & \textbf{13.1} & 50.3             & 32.1                                 & 40.9                    \\
		& D2-Net\cite{narayan2021d2}, ICCV2021        & 65.7          & 60.2          & 52.3          & 43.4          & \textbf{36.0}          & -             & -             & 51.5             & -                                    & -                       \\
		& FAC-Net\cite{huang2021foreground}, ICCV2021       & 67.6          & 62.1          & 52.6          & 44.3          & 33.4          & 22.5          & 12.7          & 52.0             & 33.1                                 & 42.2                    \\
		& ACM-Net\cite{qu_2021_acmnet}, arXiv2021       & 68.9          & 62.7          & 55.0          & 44.6          & 34.6          & 21.8          & 10.8          & 53.2             & 33.4                                 & 42.6                    \\ \cline{2-12}
		& \textbf{FTCL(Ours)} & \textbf{69.6} & \textbf{63.4} & \textbf{55.2} & \textbf{45.2} & 35.6 & \textbf{23.7} & 12.2          & \textbf{53.8}    & \textbf{34.4}                        & \textbf{43.6}           \\ \bottomrule
	\end{tabular}
}
	\vspace{-3mm}
\end{table*}

\subsection{Experimental Setup}
\noindent\textbf{THUMOS14.} It contains 200 validation videos and 213 test videos annotated with temporal action boundaries from 20 action categories. Each video contains 15.4 action instances on average, making this dataset challenging for weakly-supervised temporal action localization. Following previews works \cite{liu2021blessings,zhang2021cola,qu_2021_acmnet,yang2021uncertainty,hong2021cross}, we apply the validation set for training and the test set for evaluation.

\noindent\textbf{ActivityNet1.3.} ActivityNet1.3 contains 10,024 training videos and 4,926 validation videos from 200 action categories, and each video contains 1.6 action instances on average. 
Following the standard protocol in previous work \cite{liu2021blessings,zhang2021cola,qu_2021_acmnet,yang2021uncertainty,hong2021cross}, we train on the training set and test on the validation set.

\noindent\textbf{Evaluation Metrics.}  Following previous models~\cite{liu2021weakly,paul2018wtalc,wang2017untrimmednets}, we use mean Average Precision (mAP) under different temporal Intersection over Union (t-IoU) thresholds as evaluation metrics. The t-IoU thresholds for THUMOS14 is [0.1:0.1:0.7] and for ActivityNet is [0.5:0.05:0.95].

\noindent\textbf{Implementation Details.} Following existing methods, we use I3D \cite{carreira2017quo} model pretrained on Kinetics dataset as the RGB and optical flow feature extractors. 
The dimension of the output feature is 2048. Note that no fine-tuning operations are applied to the I3D feature extractor for a fair comparison. The number of sampled snippets $T$ for THUMOS14 and ActivityNet is set to 750 and 75, respectively. 
{To implement $f_\alpha(\cdot)$ and $f_{cls}(\cdot)$, we adopt the pre-trained ACM-Net~\cite{qu_2021_acmnet} as the backbone for video-level classification. 
For FSD contrasting, we select action/background proposals 
by using the learned CAS. For LCS contrasting, to save the computational cost, we do not use the entire untrimmed video but select the top-$J$ activated snippets for contrasting, $J$ is set to 30 and 10 for THUMOS14 and ActivityNet, respectively. The output dimension of $f_\mu(\cdot)$ and $f_g(\cdot)$ is 1024. For simplicity, $f_h(\cdot)$ is the same with $f_g(\cdot)$.
The temperature hyper-parameter $\gamma$ and threshold $\tau$  in~\eqnref{smoothMax} and~\eqnref{lcs} are 10 and 0.92. 
Our model is implemented with PyTorch 1.9.0, and we utilize Adam with a learning rate of $ 10^{-4} $ and a batch size of 16 for optimization. We train our model until the training loss is smooth. 

\vspace{\subsecmargin}
\subsection{Comparison with State-of-the-art Methods}
\vspace{\subsecmargin}

\noindent\textbf{Evaluation on THUMOS14.}
As shown in ~\tabref{THUMOS}, FTCL outperforms previous weakly supervised methods in almost all IoU metrics on the THUMOS14 dataset. Specifically, our method achieves favorable performance of $35.6\%$ mAP@0.5 and $43.6\%$ mAP@Avg. And an absolute gain of $1.4\%$ and $1.0\%$ is obtained in terms of the average mAP when compared to the SOTA approaches ACM-Net~\cite{qu_2021_acmnet} and FAC-Net~\cite{huang2021foreground}. Furthermore, we observe that our methods can even achieve comparable performance with several fully-supervised methods, although we utilize much less supervision during training. Note that CoLA~\cite{zhang2021cola} gets a higher mAP@0.7 than ours. However, we get $2.7\%$ absolute gains at average mAP. CoLA adopts a hard snippet mining strategy to pursue action completeness, which can be further equipped with our FTCL for more effective WSAL.

\noindent\textbf{Evaluation on ActivityNet1.3.} As in ~\tabref{ActivityNet3}, our method also achieves state-of-the-art performance on the ActivityNet1.3 datasets. Specifically, compared with state-of-the-art ACM-Net\cite{qu_2021_acmnet}, we obtain the relative gain of $0.8\%$. Note that the performance improvement on this dataset is not as significant as it on the THUMOS14 dataset; the reason may lie in that videos in ActivityNet are much shorter than those in THUMOS14. ActivityNet only contains 1.6 instances per video on average, while the number in THUMOS14 is 15.6. Obviously, sufficient temporal information can facilitate the fine-grained temporal contrasting.

\begin{table}[]
\centering
\caption{Comparison results on ActivityNet1.3 dataset.}
\vspace{-3mm}
\label{tab:ActivityNet3}
\resizebox{65mm}{!}{%
	\renewcommand{\arraystretch}{1.1} 
\begin{tabular}{c|cccc}
\hline
\multirow{2}{*}{Method} & \multicolumn{4}{c}{mAP@t-IoU(\%)}                                      \\ \cline{2-5} 
                        & 0.5           & 0.75          & 0.95         & \multicolumn{1}{l}{Avg} \\ \hline
STPN\cite{nguyen2018weakly}, CVPR2018          & 26.3          & 16.9          & 2.6         & 16.3                    \\
MAAN\cite{yuan2019marginalized}, ICLR2019          & 33.7          & 21.9          & 5.5          & -                       \\
Bas-Net\cite{lee2020background}, AAAI2020       & 34.5          & 22.5          & 4.9          & 22.2                    \\
A2CL-PT\cite{min2020adversarial}, ECCV2020       & 36.8          & 22.0          & 5.2          & 22.5                    \\
Lee et al\cite{lee2021learningSP}, AAAI2021    & 37.0          & 23.9          & 5.7          & 23.7                    \\
FAC-Net\cite{huang2021foreground}, ICCV2021       & 37.6          & 24.2 & 6.0          & 24.0                    \\
ACM-Net\cite{qu_2021_acmnet}, arXiv2021           & \textbf{40.1} & 24.2 & 6.2          & 24.6                    \\ \hline
\textbf{FTCL(Ours)} & 40.0          & \textbf{24.3} & \textbf{6.4} & \textbf{24.8}           \\ \hline
\end{tabular}
}
\vspace{-3mm}
\end{table}


\vspace{\subsecmargin}
\subsection{Further Remarks}\label{sec:ablation}
\vspace{\subsecmargin}
\vspace{-1mm}
To better understand our algorithm, we conduct ablation studies and in-depth analysis on the THUMOS14 dataset.

\noindent\textbf{Effectiveness of FSD Contrasting.}  We utilize FSD contrasting for discriminative foreground-background separation. To evaluate the effectiveness of this contrasting, we wipe out this module (denoted as FTCL(w/o FSD)) from the full model and observe a significant decrease in performance, as shown in~\tabref{ablation_1}. Specifically, our full model FTCL outperforms the baseline by relative gains of ($0.8\%$, $1.7\%$, $2.9\%$, $6.1\%$) mAP on t-IoU thresholds of [0.10, 0.30, 0.50, 0.70]. Without the FSD contrasting, fine-grained foreground-background distinctions can not be well handled, leading to insufficient classifier learning.

\noindent\textbf{Effectiveness of LCS Contrasting.}  We also remove LCS contrasting from the full model 
(FTCL(w/o LCS)) to evaluate its contribution to the overall performance, and the corresponding performance consistently drops as shown in~\tabref{ablation_1}, proving the positive impact for robust classification-to-localization adaption. Mining LCS for untrimmed videos enables the model to discover coherent snippets in an action instance, thus facilitating localization performance.

\noindent\textbf{Are the Above Two Modules Redundant?} Both the FSD and LCS objectives are adopted for sequence-to-sequence contrasting but with different goals. Astute readers may be curious about whether the FSD and LCS learning strategies are redundant, \ie, can we adopt either FSD or LCS for jointly modeling the foreground-background separation and classification-to-localization adaption? To answer this question, we conduct experiments with only FSD or LCS contrasting for tackling both the separation and adaption objectives, namely FTCL(both-FSD) and FTCL(both-LCS) in~\tabref{ablation_1}. We observe that our full model outperforms both variants, proving that the above two modules are not redundant. Another observation is that the two variants achieve better performance than FTCL(w/o FSD)) and FTCL(w/o LCS)). The reason lies in that both FSD and LCS belong to the sequence-to-sequence measurement, which can promote the separation and adaption objectives solely. However, since the two objectives have their unique properties, we design the FSD and LCS contrasting strategies to address them, which obtains the best performance.

\begin{table}[]
\centering
\caption{Ablation study of module effectiveness on THUMOS14.}
\vspace{-3mm}
\label{tab:ablation_1}
\resizebox{65mm}{!}{%
\renewcommand{\arraystretch}{1.1} 
\begin{tabular}{@{}cccccc@{}}
\toprule
\multicolumn{1}{c|}{}               & \multicolumn{5}{c}{mAP@t-IoU(\%)}                                                                                \\
\multicolumn{1}{c|}{}               & 0.1                  & 0.3                  & 0.5                  & 0.7                  & Avg                  \\ \midrule
\multicolumn{1}{c|}{ACM-Net}  & 68.9                 & 55.0                 & 34.6                 & 10.8                 & 42.6                 \\
\multicolumn{1}{c|}{FTCL(w/o FSD)}  & 69.0                 & 54.3                 & 34.6                 & 11.5                 & 42.8                 \\
\multicolumn{1}{c|}{FTCL(w/o LCS)}  & 69.3                 & 55.0                 & 34.8                 & 11.4                 & 43.0                 \\
\multicolumn{1}{c|}{FTCL(both-FSD)} & 69.6                 & 55.0                 & 35.3                 & 11.8                 & 43.2                 \\
\multicolumn{1}{c|}{FTCL(both-LCS)} & 69.4                 & 55.1                 & 34.8                 & 11.5                 & 43.1                 \\ \midrule
\multicolumn{1}{c|}{FTCL}           & \textbf{69.6}        & \textbf{55.2}        & \textbf{35.6}        & \textbf{12.2}        & \textbf{43.6}        \\ \midrule
\multicolumn{1}{l}{}                & \multicolumn{1}{l}{} & \multicolumn{1}{l}{} & \multicolumn{1}{l}{} & \multicolumn{1}{l}{} & \multicolumn{1}{l}{}
\end{tabular}
}
\vspace{-8mm}
\end{table}

\noindent\textbf{Why Not Resort to other Dynamic Programming Strategies like DTW?} We observe that some recent works are pursuing the video sequence alignment based on dynamic time warping (DTW)~\cite{hadji2021representation,cao2020few,dvornik2021drop}. However, DTW assumes that the two sequences can be fully aligned, thus requiring trimmed videos. To validate the effectiveness of our FTCL, as shown in~\tabref{ablation_2}, we compare our proposed method with the current state-of-the-art DTW-based approaches, Cycle-Consistency DTW (CC-DTW)~\cite{hadji2021representation} and  Drop-DTW~\cite{dvornik2021drop}. The results consistently demonstrate the superiority of our framework. We also replace our FSD and LCS strategies (\eqnref{fsd} and \eqnref{lcs}) with the standard differential DTW operator~\cite{hadji2021representation} (denoted as DTW), which obtains inferior results as we analyzed above.

\begin{table}[]
\centering
\caption{Comparison with DTW-based methods on THUMOS14.}
\vspace{-3mm}
\label{tab:ablation_2}
\resizebox{65mm}{!}{%
	\renewcommand{\arraystretch}{1.1} 
\begin{tabular}{@{}c|ccccc@{}}
\toprule
         & \multicolumn{5}{c}{mAP@t-IoU(\%)}                                             \\
         & 0.1           & 0.3           & 0.5           & 0.7           & Avg           \\ \midrule
CC-DTW\cite{hadji2021representation}   & 69.1          & 54.9          & 34.8          & 11.2          & 42.9          \\
Drop-DTW\cite{dvornik2021drop} & 69.5          & 55.2          & 35.4          & 11.3          & 43.2          \\
DTW\cite{hadji2021representation}      & 69.2          & 55.1          & 35.0          & 11.7          & 43.1          \\ \midrule
FTCL     & \textbf{69.6} & \textbf{55.2} & \textbf{35.6} & \textbf{12.2} & \textbf{43.6} \\ \bottomrule
\end{tabular}
}
	\vspace{-2mm}
\end{table}

\noindent\textbf{Complementary Role of the proposed FTCL.} It is obvious that the proposed strategy is model-agnostic and non-intrusive, and hence can play a complementary role over existing methods. In~\tabref{ablation_3}, we plug our FSD and LCS contrasting into three WSAL approaches including STPN~\cite{nguyen2018weakly}, W-TALC~\cite{paul2018wtalc}, and CoLA~\cite{zhang2021cola}. The results show that our proposed learning strategies can consistently improve their performance. In addition, our method does not introduce computational cost during model inference. Note that CoLA also adopts contrastive learning in snippet-level, while our proposed method can further boost its performance by additionally considering the fine-grained temporal distinctions.


\begin{table}[]
	\centering
	\caption{Evaluation of the complementary role of FTCL.}
	\vspace{-3mm}
	\label{tab:ablation_3}
	\resizebox{80mm}{!}{%
		\renewcommand{\arraystretch}{1.1} 
		\begin{tabular}{@{}c|ccccc@{}}
			\toprule
			& \multicolumn{5}{c}{mAP@t-IoU(\%)} \\
			& 0.1   & 0.3  & 0.5  & 0.7  & Avg  \\ \midrule
			STPN\cite{nguyen2018weakly}(reproduced)    & 52.2      & 35.6     & 16.8     & 4.1      &  27.2    \\
			\textbf{STPN+FTCL}    & \textbf{54.1(+1.9)}      &  \textbf{38.4(+2.8)}    &  \textbf{18.2(+1.4)}    &   \textbf{4.8(+0.7)}   &  \textbf{29.0(+1.8)}    \\  \midrule
			W-TALC\cite{paul2018wtalc}(reproduced)  &  55.7     &   40.0   &  22.7    &   7.7   &   34.5   \\
			\textbf{W-TALC+FTCL}  &  \textbf{57.5(+1.8)}     &  \textbf{40.9(+0.9)}    &  \textbf{23.8(+1.1)}    &  \textbf{8.4(+0.7)}    &   \textbf{35.7(+1.2)}   \\  \midrule
			CoLA\cite{zhang2021cola}(reproduced)    &  66.1     &  52.1    & 34.3     &  13.1    &  41.7    \\
			\textbf{CoLA+FTCL}    &  \textbf{67.1(+1.0)}     &  \textbf{52.9(+0.8)}    &   \textbf{34.8(+0.5)}   &   \textbf{13.2(+0.1)}   &   \textbf{42.3(+0.6)}   \\ \bottomrule
		\end{tabular}
	}
	\vspace{-4mm}
\end{table}

%


\vspace{\secmargin}
\vspace{-1mm}
\section{Conclusions}\label{sec:conclusion}
\vspace{\secmargin}
\vspace{-1mm}
This paper proposes a fine-grained temporal contrastive learning framework for WSAL, which jointly enjoys the merits of discriminative action-background separation and alleviated task gap between classification and localization. Specifically, two types of contrasting strategies, including FSD and LCS contrasting, are designed via differentiable dynamic programming,  capable of making fine-grained temporal distinctions. The encouraging performance is demonstrated in extensive experiments.

\noindent\textbf{Limitations.} In this work, similar to existing WSAL models, we equally employ a fixed snippet division strategy for all videos. However, since different videos have different duration and shots, the simple and fixed way may hinder the fine-grained temporal contrastive learning. In the future, we plan to conduct FTCL in an adaptive manner, \eg, considering hierarchical temporal structures or performing shot detection and action localization in a unified framework.

\vspace{-2mm}
\section*{Acknowledgements}
\vspace{-2mm}
This work was supported by the National Key Research \& Development Plan of China under Grant 2020AAA0106200, in part by the National Natural Science Foundation of China under Grants 62036012, 61721004, U21B2044, 62102415, 62072286, 61720106006, 61832002, 62072455, 62002355, and U1836220, in part by the Key Research Program of Frontier Sciences of CAS under Grant QYZDJSSWJSC039, in part by Beijing Natural Science Foundation (L201001), in part by Open Research Projects of Zhejiang Lab (NO.2022RC0AB02).
\clearpage
{\small
\bibliographystyle{ieee_fullname}
\bibliography{cvpr2022bib}
}

\end{document}